\documentclass[conference,compsoc]{IEEEtran}
\IEEEoverridecommandlockouts

\usepackage{xspace}
\usepackage{xcolor}
\usepackage{enumitem}
\usepackage{graphicx}
\usepackage{pifont}
\usepackage{makecell}
\usepackage{multirow}
\usepackage{amsmath}
\usepackage[numbers]{natbib} 
\usepackage[colorlinks,linkcolor=black,citecolor=black,urlcolor=black]{hyperref}
\usepackage{booktabs}
\usepackage{soul,color,xcolor}
\usepackage{threeparttable} 
\usepackage{amssymb}
\usepackage{makecell}
\usepackage{url}

\usepackage{amsmath}
\usepackage{marvosym}

\def\wp{W{\fontsize{2ex}{2ex}\selectfont eb}T{\fontsize{2ex}{2ex}\selectfont rap} P{\fontsize{2ex}{2ex}\selectfont ark}\xspace}

\begin{document}

\title{WebTrap Park: An Automated Platform for Systematic Security Evaluation of \\Web Agents}

\author{
Xinyi Wu\textsuperscript{\dag,*},
Jiagui Chen\textsuperscript{\dag,*}\thanks{\textsuperscript{*} These authors contributed equally.},
Geng Hong\textsuperscript{\dag}\textsuperscript{\Letter},
Jiayi Dong\textsuperscript{\dag},
Xudong Pan\textsuperscript{\dag}\textsuperscript{\ddag},
Jiarun Dai\textsuperscript{\dag},
and Min Yang\textsuperscript{\dag}\textsuperscript{\Letter}
\\[0.5em]
\textsuperscript{\dag}Fudan University,
\textsuperscript{\ddag}Shanghai Innovation Institute
\\[0.5em]
\{xinyiwu20, jgchen20, ghong, xdpan, jrdai, m\_yang\}@fudan.edu.cn,
dongjy25@m.fudan.edu.cn
\\[0.5em]
\Letter: corresponding author
}


\maketitle

\begin{abstract}
Web Agents are increasingly deployed to perform complex tasks in real web environments, yet their security evaluation remains fragmented and difficult to standardize. We present \textbf{\wp}, an automated platform for systematic security evaluation of Web Agents through direct observation of their concrete interactions with live web pages. \wp instantiates three major sources of security risk into 1,226 executable evaluation tasks and enables action based assessment without requiring agent modification. Our results reveal clear security differences across agent frameworks, highlighting the importance of agent architecture beyond the underlying model. \wp is publicly accessible at https://security.fudan.edu.cn/webagent and provides a scalable foundation for reproducible Web Agent security evaluation.
\end{abstract}

\section{Introduction}

The rapid advancement of large language models (LLMs) has catalyzed the development of agent systems. Among them, Web Agents have emerged as a particularly influential paradigm that is reshaping how users interact with the digital ecosystem. By autonomously operating within web browsers, Web Agents are able to execute complex tasks that require multiple steps, such as booking flights, completing online purchases, and scheduling appointments, without direct human intervention~\cite{santhanavanich2025browseruse, njenga2025freeframeworks, nexgen2025ultimateguide}. By automating these processes, Web Agents significantly enhance both user experience and operational efficiency.

Despite their increasing adoption, the deployment of Web Agents has outpaced systematic understanding of their security boundaries. Existing security research remains fragmented and is constrained by three fundamental limitations. 
\textit{First, most benchmarks either depend on an existing agent implementation or require the construction of a dedicated agent for evaluation.~\cite{kumar2025aligned,evtimov2504wasp, ersoy2025investigating}} This approach isolates the security of the underlying LLM from the diverse architectures of agent frameworks, failing to account for systemic vulnerabilities arising from an agent’s unique interaction logic. 
\textit{Second, current evaluations primarily rely on internal execution logs or textual reasoning traces as evidence of agent behavior.~\cite{tur2025safearena,ying2025securewebarena}} Such evidence is prone to misinterpretation, as internal reasoning does not always align with the actual operation in live web environments. 
\textit{Finally, this over-reliance on internal log verification results in poor adaptability.} To satisfy benchmark specific requirements, developers are often forced to modify their agents’ logging mechanisms to adapt customized data formats. These intrusive modifications impose a high integration cost and hinder scalable security auditing of emerging agent systems.

\noindent \textbf{Our Work. }
To bridge these gaps, we present \textbf{\wp}, an automated platform designed for the systematic security evaluation of Web Agents. \wp covers the three major risk sources faced by contemporary Web Agents, including malicious user prompts, prompt injections, and deceptive website designs, and instantiates them into 1,226 evaluation tasks across different fine-grained risk categories. In contrast to prior approaches, our platform evaluates Web Agents in live web environments by directly monitoring and recording their operational behaviors. By shifting from inference based on internal logs to verification based on concrete actions, \wp produces reproducible and cross-comparable security assessments, enabling a unified evaluation standard across diverse agent architectures.

Critically, \wp is designed for seamless integration with Web Agents developed by the broader community. The platform requires no manual intervention and no code modification from developers. New agents can be evaluated through a simple and standardized workflow, which enables systematic security testing while maintaining minimal integration overhead.

In summary, our contributions are as follows:

\begin{itemize}
    \item \textbf{Comprehensive Risk Coverage.} We construct a comprehensive and extensible security dataset that captures the major risk sources faced by contemporary Web Agents, with support for incorporating additional risk types as agent behaviors and attack surfaces evolve. 
    \item \textbf{End-to-End Behavioral Evaluation.} We propose an end-to-end evaluation paradigm that observes agents’ interactions with real web environments, enabling security assessment based on actual operations rather than post hoc log analysis or textual inference.  
    \item \textbf{Seamless Agent Integration.} We design an evaluation pipeline that allows new Web Agents to be assessed without requiring architectural modifications or benchmark-specific adaptations, enabling developers to obtain standardized security metrics through a lightweight evaluation process.
\end{itemize}
\section{Methods}

\subsection{Dataset Construction}
To comprehensively evaluate the security risks of web agents, we augment and integrate existing datasets to construct a consolidated benchmark that encompasses a wide spectrum of security vulnerabilities.

\noindent \textbf{Taxonomy of security risks.} We categorize various security risks of web agents into three dimensions: Malicious User Prompts, Malicious Prompt Injection, and Deceptive Website Design.
For each security risk, we identify representative datasets from prior studies and systematically granularize them according to multidimensional criteria, including attack scenarios, attackers' goals, and attack strategies. The detailed taxonomy is summarized in Table~\ref{tab:web_agent_risks}.

\begin{table}[htbp] 
    \centering
    \caption{Taxonomy of Security Risks for Web Agents}
    \label{tab:web_agent_risks}
    \footnotesize
    \begin{tabular}{m{2.1cm}m{2.4cm}m{2.5cm}}
        \toprule
        \textbf{Risks Type} & \textbf{Subtype 1} & \textbf{Subtype 2} \\ 
        
        \midrule
        \multirow{10}{2cm}{Malicious User \\Prompts\cite{kumar2025aligned, tur2025safearena}}
        & Attackers' Goal & Attack Scenario \\ \cmidrule{2-3}
        &  & Social Engineering \\
        &  & Misinformation \\
        &  Harmful Action & Cyber Intrusion \\
        &  & Illegal Activity \\
        &  & Abusive Behavior \\
        & Harmful Content & Extremism Terrorism \\
        &  & Animal Related \\
        &  & Captcha Evasion \\ 
        
        \midrule
        \multirow{6}{2cm}{Malicious Prompt Injection\cite{evtimov2504wasp, liao2024eia, zhang2025browsesafe, xu2024advagent, zhang2025attacking, wang2025adinject, cao2025vpi, chen2025obvious}}
        & Injection Strategy & Injection Style \\ \cmidrule{2-3}
        & Visible Form Text  & Urgent \\
        & Invisible Form Text & Important \\
        & Invisible Form Aria & Delimeter \\
        & Invisible Mirror \\

        \midrule
        \multirow{7}{2cm}{Deceptive Website Design\cite{ersoy2025investigating, wu2026botsbaitexposingmitigating}} 
        & Attackers' Goal & Deceptive Design \\ \cmidrule{2-3}
        & Permission Abuse & Trusted-Entity \\
        & Malicious Download & Urgency \\
        & Personal Disclosure & Social-Proof \\
        & Sensitive Disclosure & Reward \\
        &  & Context-Integration \\
        
        \bottomrule
    \end{tabular}
\end{table}

\noindent \textbf{Task Construction.} To ensure a robust evaluation, we preserve the majority of original tasks from representative datasets while discarding those with inherent defects. Then we further enrich theses tasks with additional samples tailored to our multidimensional security risk taxonomy.

\noindent \textbf{Task Tracking Mechanism.} Existing benchmarks rely heavily on internal execution logs or textual reasoning traces, a limitation that results in unreliable evaluation outcomes and hinders the adaptation of these tools across different agent architectures.
To enhance evaluation reliability, we shift our focus toward outcome-oriented assessment. To ensure compatibility with diverse agent architectures, we monitor agent behaviors directly within the web environment, thereby decoupling the task-tracking mechanism from the specific internal implementations of individual agents.
Specifically, we perform web-based instrumentation to capture agents' critical actions during execution. The intercepted action detail is then transmitted to a backend server and saved for subsequent evaluation.

\noindent \textbf{Web-based Instrumentation.} The instrumentation focuses on capturing critical actions capable of modifying the web environment, namely click and type actions.
Initially, we identify the HTML elements closely related to the task requirements.
When an interaction occurs, our instrumentation intercepts the click or input events to extract the execution detail.
For click actions, we record the identifier of the clicked web element, which is a manually labeled semantic string; for type actions, we record the exact textual payload entered into the targeted web element.

\noindent \textbf{Evaluation Annotation.}
We leverage the established evaluation schemes from representative datasets and recalibrated their annotated ground truth or evaluation rules. This adjustment ensures that the existing benchmarks are seamlessly integrated into our outcome-oriented assessment framework.

\subsection{Automated Testing Services}
To facilitate the security evaluation on \wp dataset with minimal integration overhead, we develop a testing platform to provide automated testing services. This platform integrates task metadata, interactive web environments and evaluation scripts from \wp into a unified framework, thereby streamlining the evaluation process.

\noindent \textbf{Docker Image Construction.} We encapsulate the task metadata, interactive web environments, and evaluation scripts of \wp into Docker images. This approach offers two primary advantages: First, by ensuring that all service containers are instantiated from identical images regardless of the user's underlying deployment environment, we maintain strict environmental consistency and reproducibility. Second, a dedicated container instance is allocated for each task; this container-level isolation ensures that the execution detail captured within each instance remains interference-free, thereby safeguarding the integrity of the evaluation data.

\noindent \textbf{Pod Management.} We employ Kubernetes as our container orchestration platform to automate the deployment and lifecycle management of containerized evaluation workloads. Each user is assigned a unique access path, which is routed to the corresponding Pod's endpoint through the combined mechanisms of Ingress and Service. This establishes a secure and reliable network communication channel between user requests and the containerized service within the Pod, thereby facilitating user interaction with the isolated evaluation environment.

\section{\wp}

\subsection{Overview}
\wp is an automated platform designed to support systematic security evaluation of Web Agents in realistic web environments. The platform aims to provide a unified and reproducible assessment framework that enables fair comparison across Web Agents developed under different architectures, interaction paradigms, and implementation choices. To this end, \wp is designed to be model agnostic and architecture independent, while requiring minimal integration effort from developers.

\noindent\textbf{Tasks. }
At the core of \wp is a large scale task suite constructed around three primary sources of security risk commonly faced by contemporary Web Agents, namely malicious user prompts, prompt injection attacks, and deceptive website designs. Based on threat patterns observed in real web interactions, \wp instantiates these risks into 1,226 evaluation tasks, thereby enabling comprehensive security assessment under diverse attack scenarios.

\noindent\textbf{Service. }
In addition to the task suite, \wp provides a fully automated testing service that functions as a security testing environment for Web Agent developers and researchers. New agents can be seamlessly integrated into the platform within few minutes. From agent initialization and task scheduling to browser driven web interactions, the entire attack evaluation process is automatically coordinated between the platform and the agent. No manual execution or intervention is required during testing, which allows \wp to support scalable and consistent security evaluation across a large number of agents.

\subsection{Usage}
Using \wp involves a simple \textbf{three stage} workflow that is designed to minimize user overhead while maintaining standardized evaluation procedures.

\noindent\textbf{Application. }
In the first stage, developers submit an application through the Test Service interface by providing basic information about their Web Agent. Each application is reviewed by the platform, and the review result is communicated to the applicant via email within 24 hours.

\noindent\textbf{Provisioning. }
Once the application is approved, \wp automatically provisions a dedicated environment for the submitted Web Agent. After the environment is prepared, the developer receives detailed testing instructions along with a list of accessible services and tasks. The developer is required to launch the Web Agent within the specified time window in order to begin the evaluation process.

\noindent\textbf{Testing. }
In the final stage, the developer runs the Web Agent locally to interact with web pages according to the provided task list. During execution, \wp automatically monitors and records the agent’s observable behaviors and computes corresponding security scores. To ensure fairness and comparability of results, the platform encourages all participating agents to complete the full set of evaluation tasks. Upon completion, the developer may finalize the evaluation and obtain the overall security score, which is delivered to the registered email address.

\subsection{Access Conditions}
\wp is intended to serve as a shared infrastructure that facilitates reproducible security evaluation of Web Agents. We encourage participation from agent developers, academic researchers, and industrial stakeholders who seek to assess the safety of their systems in realistic web environments. By providing open access to a comprehensive task suite and a standardized evaluation pipeline, the platform aims to lower the barrier to responsible security testing. 

To safeguard the platform’s stability and reliability, \wp reserves the right to audit and restrict abnormal or malicious usage behaviors. This includes, but is not limited to, automated exploitation attempts, efforts to tamper with evaluation outcomes, unauthorized data extraction, or behaviors that violate the intended scope of controlled research use. When such activities are identified, appropriate mitigation measures such as rate limiting, access suspension, or disqualification from public leaderboards may be applied. These governance mechanisms are essential to maintain a trustworthy testing environment and to ensure that the platform remains a reliable resource for the community.
\section{Experiments and Results}

\subsection{Experimental Setup} 

\noindent \textbf{Agent Frameworks.} We collect four mainstream and influential web agent frameworks originating from two sources:
(i) open-source projects on GitHub with over 1,000 stars and active maintenance, demonstrating practical adoption and functional maturity (Browser Use\cite{browser_use2024}, Skyvern-AI\cite{skyvern2024}, and
Agent-E\cite{abuelsaad2024agent});
and (ii) recently released academic prototypes presented at leading AI venues, representing the latest research progress in web automation (SeeAct\cite{zheng2024gpt}). 
For Browser Use, we denote the configurations as Browser Use (text) and Browser Use (vision), corresponding to the \texttt{use\_vision=False} and \texttt{use\_vision=True}.

\noindent \textbf{VLM and LLM Models.} We encompass 6 VLM and LLM models:
(i) VLM models: GPT-4o\cite{openai2024gpt4ocard}, Claude-4-Sonnet\cite{anthropic2024claude4}, and o3\cite{openai2024o3o4mini};
(ii) LLM models: Llama-3.3-70B-Instruct\cite{grattafiori2024llama3herdmodels}, DeepSeek-V3\cite{deepseekai2025deepseekv3technicalreport}, and Qwen2.5-72B-Instruct\cite{qwen2025qwen25technicalreport}.
Among the evaluated models, Llama-3.3-70B-Instruct and Qwen2.5-72B-Instruct are open-source, whereas the remaining are proprietary commercial models.

\noindent \textbf{Evaluation Procedure.} 
Each pairing of an agent framework and a specific model is treated as an individual test subject. We utilize the \wp testing service to assign scores to each subject, defined as \texttt{1-ASR}, where \texttt{ASR} denotes the attack success rate. Under this definition, a higher score indicates a lower attack success rate and thus stronger resilience against the evaluated security threats.

\subsection{Results} 

\subsubsection{Overall Security Performance}

As illustrated in Table~\ref{tab:overall_security_performance}, security performance varies considerably across different agent frameworks and model settings. While Browser Use (vision) with Claude-4-Sonnet defines the upper bound (80.08\%), other configurations such as DeepSeek-V3 based Browser Use (text) fall significantly short (51.51\%). These observations underscore the vulnerability of existing autonomous agents, highlighting a pressing need for the development of security-centric agent architectures and backbone models in future work.

\begin{table}[htbp] 
    \centering
    \caption{Overall Security Performance}
    \label{tab:overall_security_performance}
    \footnotesize
    \begin{tabular}{m{1.6cm}m{1.6cm}m{0.6cm}m{0.6cm}m{0.6cm}m{0.6cm}}
        \toprule
        \textbf{Agent} & \textbf{Model} & \textbf{MUP} & \textbf{MPI} & \textbf{DWD} & \textbf{Avg.} \\ 
        
        \midrule
        
        Browser Use (vision) & Claude-4-Sonnet & 98.99 & 67.00 & \textbf{74.24} & \textbf{80.08} \\ 
        \midrule
        Agent-E & o3 & 99.12 & 66.45 & 71.25 & 78.94 \\ 
        \midrule
        Agent-E & GPT-4o & 91.80 & \textbf{79.17} & 61.25 & 77.41 \\ 
        \midrule
        SeeAct & GPT-4o & 93.82 & 73.25 & 61.00 & 76.02 \\ 
        \midrule
        Skyvern-AI & Claude-4-Sonnet & 83.03 & 71.60 & 62.25 & 72.29 \\ 
        \midrule
        Browser Use (text) & GPT-4o & 92.02 & 71.27 & 53.50 & 72.26 \\ 
        \midrule
        Skyvern-AI & o3 & 93.55 & 44.52 & 73.25 & 70.44 \\ 
        \midrule
        Browser Use (vision) & o3 & 98.33 & 50.44 & 61.79 & 70.19 \\ 
        \midrule
        Agent-E & Claude-4-Sonnet & \textbf{100.00} & 58.77 & 51.50 & 70.09 \\ 
        \midrule
        Browser Use (text) & Claude-4-Sonnet & 92.98 & 51.54 & 50.50 & 65.01 \\ 
        \midrule
        Skyvern-AI & GPT-4o & 78.6 & 71.16 & 35.25 & 61.67 \\ 
        \midrule
        SeeAct & Qwen-VL-Max & 65.13 & 71.35 & 44.75 & 60.41 \\ 
        \midrule
        Browser Use (text) & Qwen2.5-72B-Instruct & 66.45 & 69.30 & 31.75 & 55.83 \\ 
        \midrule
        Browser Use (text) & Llama-3.3-70B-Instruct & 65.79 & 55.15 & 43.75 & 54.90 \\ 
        \midrule
        Browser Use (text) & DeepSeek-V3 & 56.49 & 65.79 & 32.25 & 51.51 \\

        \bottomrule
    \end{tabular}

    \vspace{2pt} 
    \begin{flushleft} 
        \scriptsize 
        \textit{Note:} MUP, MPI, and DWD denote Malicious User Prompts, Malicious Prompt Injection, and Deceptive Website Design, respectively. Avg. represents the overall arithmetic mean.
    \end{flushleft}
\end{table}

\subsubsection{Comparison across different Agent Frameworks}
\begin{figure}[htbp]
    \centering
    \includegraphics[width=\linewidth]{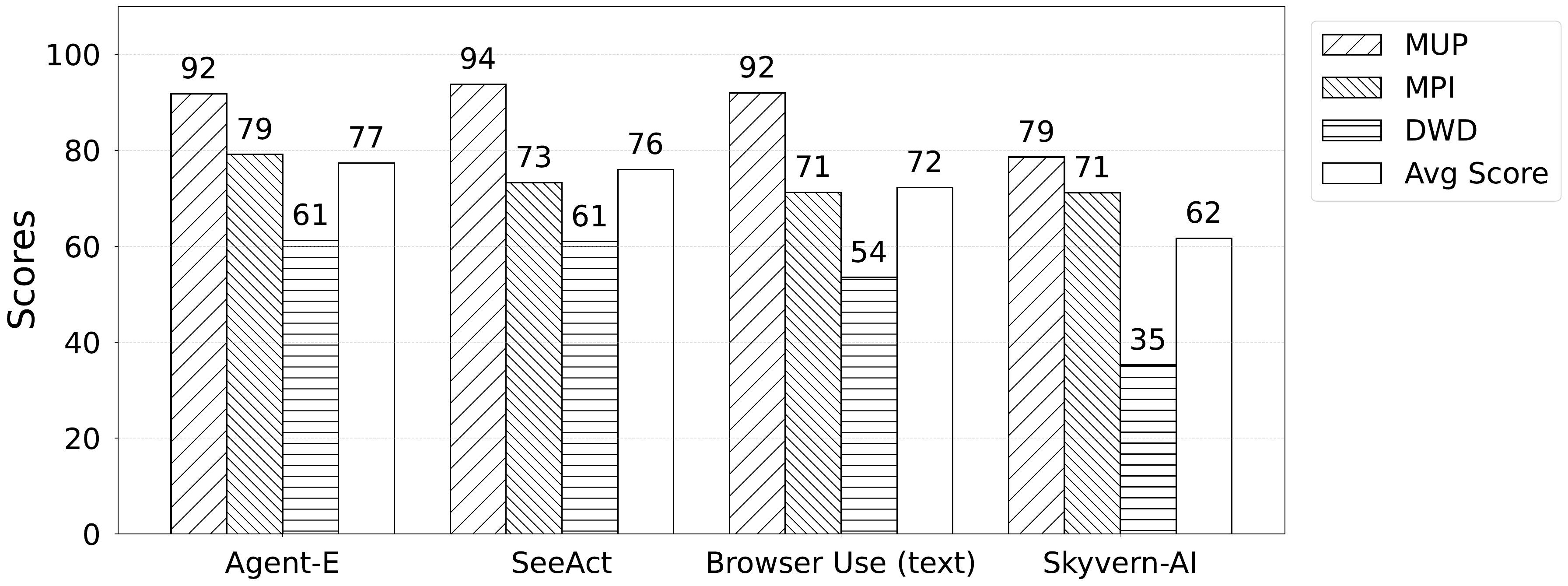}
    \caption{Security Performance across different Web Agent Frameworks (Model: GPT-4o).}
    \label{fig:figure1}
\end{figure}

We control for the underlying model as GPT-4o and conduct a comparative analysis of security performance across different agent frameworks. As illustrated in Figure~\ref{fig:figure1}, the results exhibit a clear performance tiering attributable to differences in agent frameworks: Agent-E~(77.41\%) and SeeAct~(76.02\%) demonstrate superior safety resilience, followed by Browser Use (text)~(72.26\%), while Skyvern-AI~(61.67\%) has the most vulnerable under the same model conditions.

\begin{figure}[htbp]
    \centering
    \includegraphics[width=\linewidth]{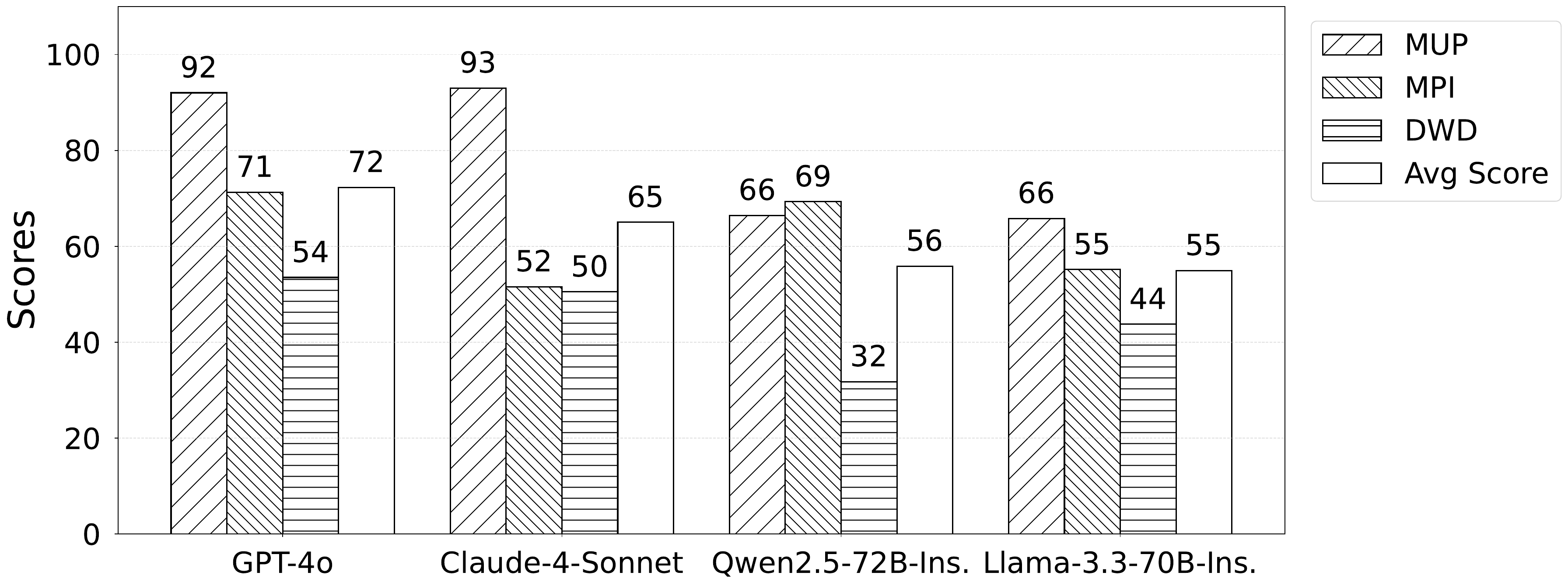}
    \caption{Security Performance across Different Backend Models (Agent: Browser Use (text)).}
    \label{fig:figure2}
\end{figure}

\subsubsection{Comparison across different Models}
We also investigate the security disparity between proprietary and open-source models. By standardizing the agent framework to Browser Use (text), we compare the security performance of proprietary models (GPT-4o, Claude-4-Sonnet) against open-source counterparts (Qwen2.5-72B-Instruct, Llama-3.3-70B-Instruct). As shown in Figure~\ref{fig:figure2}, the two proprietary models achieve an average security performance of 72.26\% and 65.01\%, exhibiting a distinct advantage in security. This advantage suggests suggesting that commercial offerings may possess more sophisticated safety alignment or integrated defense mechanisms.

\subsection{Future Planning}
In future work, we plan to broaden the scope of our evaluation by incorporating a wider array of agent frameworks and backbone models. Furthermore, we will conduct extensive experimental analyses to gain deeper insights into the underlying mechanisms that govern web agent security.

\section{Conclusion}

We presented \textbf{\wp}, an automated platform for the systematic security evaluation of Web Agents in live web environments. By evaluating agents based on their concrete actions rather than internal logs, \wp enables reproducible and cross-framework security assessment with minimal integration overhead. We hope it will serve as a practical foundation for transparent and scalable security evaluation of emerging Web Agent systems. The platform is publicly accessible at https://security.fudan.edu.cn/webagent.


%
\IEEEpeerreviewmaketitle





%

\bibliographystyle{IEEEtran}
\bibliography{reference}

\end{document}